\documentclass[]{interact}

\usepackage{fix-cm}
\usepackage{epstopdf}
\usepackage{subcaption}
\usepackage{url}
\usepackage{xcolor}
\usepackage{soul}
\renewcommand{\hl}[1]{#1}
\usepackage{hyperref}
\usepackage{multirow}
\usepackage{amsmath}
\usepackage{threeparttable}
\usepackage{optidef}
\usepackage{booktabs}
\usepackage{algorithmicx}
\usepackage[ruled]{algorithm}
\usepackage[noend]{algpseudocode}

\usepackage{amssymb}
\usepackage{pifont}

\usepackage[most]{tcolorbox}

\newcommand{\figlab}[1]{\label{fig:#1}}
\newcommand{\figref}[1]{Fig.~\ref{fig:#1}} 
\newcommand{\tablab}[1]{\label{tab:#1}}
\newcommand{\tabref}[1]{Table~\ref{tab:#1}} 

\newcommand{\seclab}[1]{\label{sec:#1}}
\newcommand{\secref}[1]{Section~\ref{sec:#1}} 
\newcommand{\subseclab}[1]{\label{subsec:#1}}
\newcommand{\subsecref}[1]{Section~\ref{subsec:#1}} 
\newcommand{\algolab}[1]{\label{algorithm:#1}}

\newcommand{\etal}{\textit{et~al.}}

\newcommand{\eg}{\textit{e.g.}}

\usepackage[numbers,sort&compress]{natbib}
\bibpunct[, ]{[}{]}{,}{n}{,}{,}
\renewcommand\bibfont{\fontsize{10}{12}\selectfont}
\makeatletter
\def\NAT@def@citea{\def\@citea{\NAT@separator}}
\makeatother

\theoremstyle{plain}

\theoremstyle{definition}

\theoremstyle{remark}

\begin{document}

\articletype{FULL PAPER}

\title{Replanning Human--Robot Collaborative Tasks with Vision--Language Models via Semantic and Physical Dual--Correction}

\author{
\name{Taichi Kato\textsuperscript{a}, Takuya Kiyokawa\textsuperscript{b*}\thanks{\textsuperscript{*}Corresponding Author: Takuya Kiyokawa. Email: kiyokawa@sys.es.osaka-u.ac.jp}, Namiko Saito\textsuperscript{b,c}, and Kensuke Harada\textsuperscript{b,d}}
\affil{\textsuperscript{a}Department of Systems Science, School of Engineering Science, The University of Osaka, 1-3 Machikaneyama, Toyonaka, Osaka, Japan; \\\textsuperscript{b}Department of Systems Innovation, Graduate School of Engineering Science, The University of Osaka, 1-3 Machikaneyama, Toyonaka, Osaka, Japan; \\\textsuperscript{c}Microsoft Research Asia, Shinagawa, Tokyo, Japan; \\\textsuperscript{d}Industrial Cyber-physical Systems Research Center, The National Institute of Advanced Industrial Science and Technology (AIST), 2-3-26 Aomi, Koto-ku, Tokyo, Japan.}
}

\maketitle

\begin{abstract}
Human--robot collaborative assembly requires robots to interpret ambiguous corrective instructions while producing physically executable motions. Vision--language models (VLMs) provide semantic reasoning but may select logically inconsistent targets or misjudge execution outcomes. \hl{We propose a replanning framework that maps human instructions to \textit{Action Target} candidates, including grasp poses and tool selections, and combines an \textit{Internal Correction Model} for pre-execution logical verification with an \textit{External Correction Model} for post-execution visual verification.} The framework integrates VLM reasoning with 6-DoF grasp generation and collision-free trajectory planning. \hl{Simulation ablations show configuration-dependent effects: internal correction improves candidate validity, whereas external correction enables recovery for a low-latency VLM but can reduce success when visual verification produces false negatives.} Experiments with an upper-body humanoid robot achieved 66.7\% success in real-world object fixation, 100\% in initial tool selection, and 75.0\% in corrective tool selection. These results demonstrate interactive replanning across spatial and semantic collaborative tasks while identifying visual-state verification as a key limitation.
\end{abstract}


\section{Introduction}
While recent advances in manufacturing industry have advanced rapidly, associated social challenges such as workforce shortages and the demand for flexible production systems have been increasing. 
To address these issues, a new concept known as Industry 5.0~\cite{EC2021Industry5} has been proposed. Unlike conventional manufacturing paradigms that focus on machine-centered automation, Industry 5.0 emphasizes human-centered production activities~\cite{coronado2022evaluating}. Within this paradigm, Human--Robot Collaboration (HRC) is regarded as a key technology for enabling adaptive and interactive assembly tasks.

Previous studies on HRC have proposed various collaborative approaches from perspectives such as motion planning~\cite{liu2024vision}, role allocation~\cite{wang2024assembly}, and safety assurance~\cite{kotake2025cooperation}. They assume predefined workflows or structured commands. In real-world assembly tasks, however, there are many situations in which humans are responsible for processes that robots cannot perform independently, while robots act in a supportive role. In such scenarios, a cooperative relationship is required in which humans take the initiative and robots provide appropriate assistance to complete the task.

One of the main challenges in human--robot collaborative assembly is enabling robots to understand human intentions and actions. When humans work together, they often use ambiguous verbal instructions, such as ``hold it a little more to the left'' or ``pass me a larger tool,'' to adjust each other's actions and smoothly coordinate the task. Humans naturally resolve such ambiguity through shared context and implicit intention, but enabling robots to do the same remains a fundamental challenge. Many existing systems rely on predefined fixed motions, making real-time fine-grained adjustments in response to human instructions difficult.

Recent research has attempted to address this problem using Vision--Language Models (VLMs), which can connect natural language understanding with visual perception. While VLMs demonstrate strong semantic reasoning capabilities, directly applying them to collaborative assembly introduces critical limitations. First, VLMs are prone to hallucinated reasoning, selecting actions that appear linguistically plausible but are logically inconsistent with the task constraints. Second, they lack awareness of physical feasibility and therefore cannot anticipate execution failures such as unstable grasps or environmental interference. In a collaborative setting, such errors are particularly problematic, as inappropriate assistance can interrupt human operation or compromise safety.

In this study, to overcome the limitations, we propose a robust replanning framework for human--robot collaborative assembly that integrates Vision--Language Models (VLMs) with a dual-correction mechanism. Our approach treats assistive actions (e.g., grasping poses or tool selections) as selectable targets and dynamically modifies them based on natural language instructions. In this framework, verbal instructions are interpreted by a VLM, which selects appropriate action candidates from a set of generated options. This enables the robot to infer human intent and adapt its behavior in real time, even from ambiguous commands.

A distinguishing feature of our framework is the integration of the dual-correction mechanism to mitigate the inherent risks of VLMs, such as hallucinations and execution failures. Specifically, the \textit{Internal Correction Model} functions as a pre-execution verifier, checking the logical consistency between the linguistic instruction and the generated action plan to prevent erroneous selection. Conversely, the \textit{External Correction Model} serves as a post-execution evaluator, comparing environmental states before and after the action to detect physical failures and trigger necessary replanning. By introducing this closed-loop correction, the system can identify the causes of hallucinations or execution failures and perform replanning accordingly, thereby improving the stability of action modification.

In our framework, we employ separate VLMs for the Internal and External Correction models. Target recognition and action success detection require a multimodal interpretation of both linguistic instructions and environmental states. Traditional rule-based methods struggle to account for the infinite variations in natural language and dynamic environmental conditions. By leveraging VLMs trained on large-scale datasets, our correction mechanism achieves semantic-level understanding that generalizes across diverse vocabularies and visual settings. This approach ensures robust performance by focusing on the underlying meaning of a task's success rather than relying on rigid, predefined rules.

Furthermore, understanding intent is insufficient without execution capability. Therefore, we seamlessly integrate the VLM-based reasoning with robust 6-DoF grasp generation and trajectory planning algorithms. This allows the system to translate abstract semantic instructions into physically executable robot commands, realizing a complete pipeline for interactive assistance in both simulation and real-world environments.

The main contributions of this work are summarized as follows:
\begin{enumerate}
  \item \hl{An \textit{Action Target} abstraction for VLM-based human--robot collaborative assembly that maps ambiguous natural language instructions to executable assistive targets, such as grasp poses and tool selections.}
  \item \hl{A semantic--physical dual-correction mechanism that combines pre-execution logical verification and post-execution physical verification to support robust replanning.}
  \item \hl{An integrated and experimentally validated system that combines VLM-based reasoning with diffusion-based 6-DoF grasp generation and collision-free trajectory planning for object-fixation and tool-preparation tasks in simulation and real-world environments.}
\end{enumerate}

\section{Related Work} \seclab{related_work}
\subsection{Collaborative Work Based on Natural Language} \subseclab{nl_collaboration}
Research on HRC has been conducted from various perspectives, including task allocation and enhanced communication.
Liu~\etal~\cite{liu2024vision} proposed an HRC assembly method enabled by an LLM. In general robotic task planning, SayCan~\cite{ahn2022do} grounds LLM-generated actions using robot affordances, while Code as Policies~\cite{liang2023code} generates robot policy code from natural language commands.
While these approaches focus on discrete task sequences, Liu~\etal~\cite{liu2024enhancing} further proposed a framework that integrates task planning from natural language instructions using LLMs with remote teaching, enabling collaborative work that includes previously unseen motions.
Furthermore, to address ambiguity in human commands, recent studies such as KnowNo~\cite{ren2023robots} have explored mechanisms for robots to quantify uncertainty and ask clarifying questions.

Other studies have explored interactive error recovery. In the work by Lim~\etal~\cite{lim2024enhancing}, humans provide task instructions to a robot, which interprets the instructions using an LLM, generates motions, and executes assembly tasks.
When an error occurs, the robot reports the error, and the human provides corrective instructions to enable re-execution.
Similarly, TidyBot~\cite{wu2023tidybot} utilizes LLMs to infer personalized preferences from user feedback to adapt robot behavior.
These studies demonstrate the effectiveness of LLMs in handling linguistic feedback and error recovery.

However, most of these systems rely on reactive human intervention when failures occur. They do not provide autonomous mechanisms to verify the logical consistency of inferred actions before execution, nor do they explicitly distinguish between semantic reasoning errors and physical execution failures. As a result, robustness in real-time collaborative assembly remains limited.

\subsection{Collaborative Work Using Vision--Language Models} \seclab{vlm_collaboration}
There are also many studies that incorporate visual information into the reasoning process.
Embodied multimodal models such as PaLM-E~\cite{driess2023palm} and RT-2~\cite{brohan2023rt} have demonstrated end-to-end capabilities from vision--language inputs to robot actions.
Mei~\etal~\cite{mei2024replanvlm} proposed a method that generates task plans and executable code from natural language instructions using a VLM, and employs an additional VLM to evaluate execution results before and after task execution.
The concept of feedback-based correction is also explored in Inner Monologue~\cite{huang2022inner} and Reflexion~\cite{shinn2024reflexion}, which improve reasoning reliability by incorporating environmental or verbal feedback.
Black~\etal~\cite{black2024pi_0} introduced a vision-language-action flow model that maps visual observations and language instructions to robot actions for general robot control.

In many of these studies, robots primarily execute tasks autonomously after receiving high-level language or vision-language instructions. Recent advancements, however, have further accelerated the integration of VLMs into smart manufacturing~\cite{fan2025vision}. Researchers have explored LLM- and VLM-enhanced multi-agent paradigms for embodied perception, decision-making, and digital twin-assisted collaborative assembly~\cite{cai2026llm,liu2026from}. Ji~\etal~\cite{ji2024foundation} proposed a framework for collaborative assembly using a foundation model that integrates natural language instructions and visual perception. Furthermore, to mitigate instruction ambiguity, recent studies have introduced seamless interactive guidance systems~\cite{liu2026ar}, vision-language-guided action planning~\cite{fan2024vision}, and frameworks such as H2R Bridge for few-shot intention meta-perception~\cite{wu2025h2r}. Additionally, VoxPoser~\cite{huang2023voxposer} uses LLMs and VLMs to compose 3D value maps for motion planning, whereas Language-to-Rewards~\cite{yu2023language} uses LLMs to generate reward functions that connect high-level instructions to low-level robot actions. By combining large language models and vision--language models, these systems perform environmental understanding and task planning, enabling flexible task support.

\hl{However, many of these frameworks mainly focus on task generation, action planning, intention perception, or completion verification, rather than modifying physically executable assistive motions in response to ambiguous human corrective instructions during assembly. In contrast to ReplanVLM}~\cite{mei2024replanvlm}\hl{, which focuses on task-plan replanning and execution-result verification, our framework introduces an \textit{Action Target} abstraction for assistive collaborative assembly, where ambiguous instructions are mapped to executable targets such as grasp poses or tool selections. This abstraction is integrated with 6-DoF grasp generation and collision-free trajectory planning, enabling VLM-based correction to be grounded into physically executable robot assistance.}

\section{Iterative Action Refinement via Semantic--Physical Verification}
\begin{figure}[t]
\centering
\includegraphics[width=1.0\linewidth]{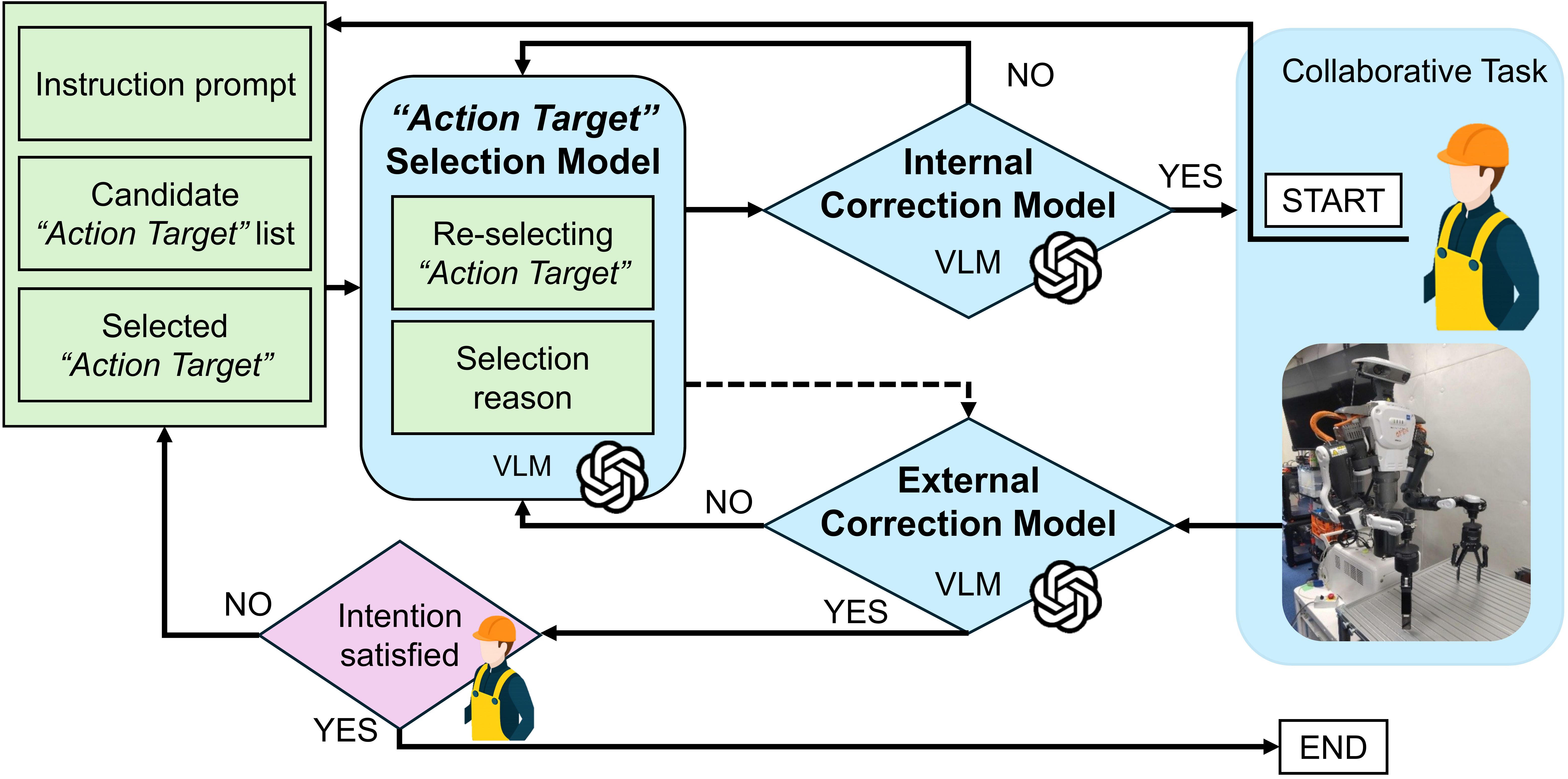}
\caption{Overview of the proposed method.}
\figlab{overview}
\end{figure}

\subsection{Overview}
This section presents a method for achieving dynamic replanning in human--robot collaborative assembly tasks leveraging VLMs. The proposed approach provides a general framework applicable to multiple types of assistive actions. To handle these diverse tasks in a unified manner, we refer to the specific subject of the robot's execution as an \textit{Action Target}. Depending on the task context, this abstraction specifically corresponds to the grasping position in object fixation, the support position in transportation, the assembly components in replenishment, or the preparatory sub-tasks in action preparation scenarios.

\figref{overview} illustrates the overall architecture and process flow.
First, the robot generates a set of \textit{Action Target} candidates, denoted as $O$, where $o_i$ represents the $i$-th candidate. These candidates are provided to a VLM together with the current environmental image $x_t$ and the persistent reasoning context $\ell_{\text{current}}$, which contains the current instruction $\ell$ and previous action history.

To ensure robustness, the selected output undergoes a dual-correction process.
Before execution, the Internal Correction Model evaluates $o_{\text{target}}$ by cross-referencing it with $x_t$ and $\ell$ to check for logical consistency (e.g., directional alignment). If an error is detected, feedback $e_{\text{logic}}$ is generated for immediate re-selection.
If valid, the robot executes the action. Subsequently, the External Correction Model verifies the result by comparing the pre-execution state $x_t$ with the post-execution state $x_{\text{new}}$. If visual failure is detected, physical feedback $e_{\text{phys}}$ is returned.

The procedural details are summarized in Algorithm~1.
The system maintains a persistent reasoning context $\ell_{\text{current}}$ throughout each collaborative task episode. The current instruction, executed \textit{Action Targets}, and their verification or correction results are appended to this context, allowing previous action history to be used in subsequent interaction turns.

\begin{algorithm}[!htbp]
\caption{Replan Collaborative Task}
\algolab{replanobject}

\footnotesize
\algrenewcommand\algorithmicindent{1em}
\algrenewcommand{\algorithmiccomment}[1]
  {\hfill{\scriptsize$\triangleright$~#1}}

\begin{algorithmic}[1]

\State Initialize $t\gets0$, $x_t$, $O$, and
$\ell_{\text{current}}\gets\emptyset$

\While{\textbf{true}}
    \State $\ell\gets\text{GetInstruction}()$
    \Comment{Human input}

    \If{$\ell$ indicates ``Task Done''}
        \State \textbf{break}
        \Comment{Finish}
    \EndIf

    \State $\ell_{\text{current}}
    \gets\ell_{\text{current}}+\ell$
    \Comment{Keep history}

    \State $n_{\text{int}}\gets0,\quad n_{\text{ext}}\gets0$
    \Comment{Retry counts}

    \While{\textbf{true}}

        \State $o_{\text{target}}
        \gets\text{VLM}_{\text{select}}
        (x_t,\ell_{\text{current}},O)$

        \If{$\text{VLM}_{\text{internal}}
        (x_t,o_{\text{target}},\ell,O)$ fails}

            \State $e_{\text{logic}}\gets\text{GetReason}()$
            \Comment{Logical feedback}

            \State $\ell_{\text{current}}
            \gets\ell_{\text{current}}+e_{\text{logic}}$

            \State $n_{\text{int}}\gets n_{\text{int}}+1$

            \State \textbf{if} $n_{\text{int}}\geq3$
            \textbf{ then return} \textsc{SelectionFailure}

            \State \textbf{continue}
        \EndIf

        \State $\ell_{\text{current}}
        \gets\ell_{\text{current}}+o_{\text{target}}$
        \Comment{Store target}

        \State Execute action on $o_{\text{target}}$

        \State $x_{\text{new}}\gets\text{CaptureImage}()$

        \State $r_{\text{ext}}
        \gets\text{VLM}_{\text{external}}
        (x_t,x_{\text{new}},\ell)$

        \State $x_{t+1}\gets x_{\text{new}},\quad t\gets t+1$
        \Comment{Update state}

        \If{$r_{\text{ext}}=\mathrm{NO}$}

            \State $e_{\text{phys}}\gets\text{GetReason}()$
            \Comment{Physical feedback}

            \State $\ell_{\text{current}}
            \gets\ell_{\text{current}}+e_{\text{phys}}$

            \State $n_{\text{ext}}\gets n_{\text{ext}}+1$

            \State \textbf{if} $n_{\text{ext}}\geq3$
            \textbf{ then return} \textsc{ExecutionFailure}

            \State \textbf{continue}
        \EndIf

        \State $\ell_{\text{current}}
        \gets\ell_{\text{current}}+\text{Success}$
        \Comment{Verified}

        \State \textbf{break}
    \EndWhile
\EndWhile

\end{algorithmic}
\end{algorithm}
\subsection{Internal Correction Model} \subseclab{innerbot}
The Internal Correction Model functions as a pre-execution semantic filter designed to prevent the execution of erroneous actions. This model validates the VLM's selected target by cross-referencing it with the current environmental image and the linguistic instruction. The validation process consists of two distinct checks. First, format validation ensures that the selected target is strictly contained within the predefined candidate set, thereby filtering out hallucinated responses or invalid formats.

Second, consistency validation evaluates the logical alignment between the selected target and the instruction within the context of the image. For instance, if the instruction directs a movement to the right, the model verifies whether the coordinate of the selected grasp pose is spatially consistent with that direction relative to the current state. If either condition is violated, the model generates descriptive feedback to trigger immediate re-selection.

To prevent infinite loops, the system terminates the task as a selection failure if the number of rejections reaches three. This specific limit was determined empirically based on extended simulation trials in the object fixation task. Our analysis of failure cases showed that allowing the correction loop to continue for four to six iterations rarely yielded successful self-correction. Instead, the selection model tended to converge, repeatedly proposing the same subset of previously rejected or ineffective grasp candidates. These observations indicate that additional retries primarily increase computation time without substantially improving the task completion probability. Thus, setting the limit to three provides a practical trade-off between allowing sufficient exploration capability and maintaining system efficiency.

\subsection{External Correction Model} \subseclab{extrabot}
The External Correction Model serves as a post-execution physical verifier to detect failures caused by external disturbances or control errors. This model receives the pre-execution image, the post-execution image, and the instruction to assess the success of the action. It compares the two images to verify whether the expected state transition, such as a tool being removed from a holder or an object's pose changing, has effectively occurred.

If the visual evidence suggests no significant change or an incorrect state, the model returns feedback explaining the physical discrepancy, allowing the VLM to update its reasoning context for the next iteration. Consistent with the Internal Correction Model, the operation is terminated after three consecutive physical failures.

\section{Experiments} \seclab{experiment}
In the experiments, we evaluated the proposed method through two representative collaborative tasks: \textit{object fixation} and \textit{tool preparation}. While the proposed framework is applicable to various assistive actions, these two tasks were selected to specifically verify the system's capability in fine-grained spatial adjustment and semantic replanning, respectively. Each task was executed in both simulation and real-world environments to validate the system's effectiveness and transferability.
\hl{The quantitative evaluation focuses on position-controlled \textit{Action Target} replanning; force adaptation is discussed only as a preliminary future extension in} \subsecref{force_universality}.

To comprehensively assess the system, we conducted three types of evaluations: (1) success rate measurement in standard tasks to validate overall performance, (2) ablation studies focusing on robustness against consecutive instructions, and (3) validation of the individual correction models using adversarial inputs to verify their detection capabilities.

\secref{sys_env} describes the experimental environments. \secref{tasks_prompts} details the tasks and prompt designs. Finally, \secref{results} reports the results.

\subsection{Experiment Platform} \seclab{sys_env}
\begin{figure*}[tb]
    \centering
    \begin{minipage}[tb]{0.57\linewidth}
        \centering
        \includegraphics[width=\linewidth]{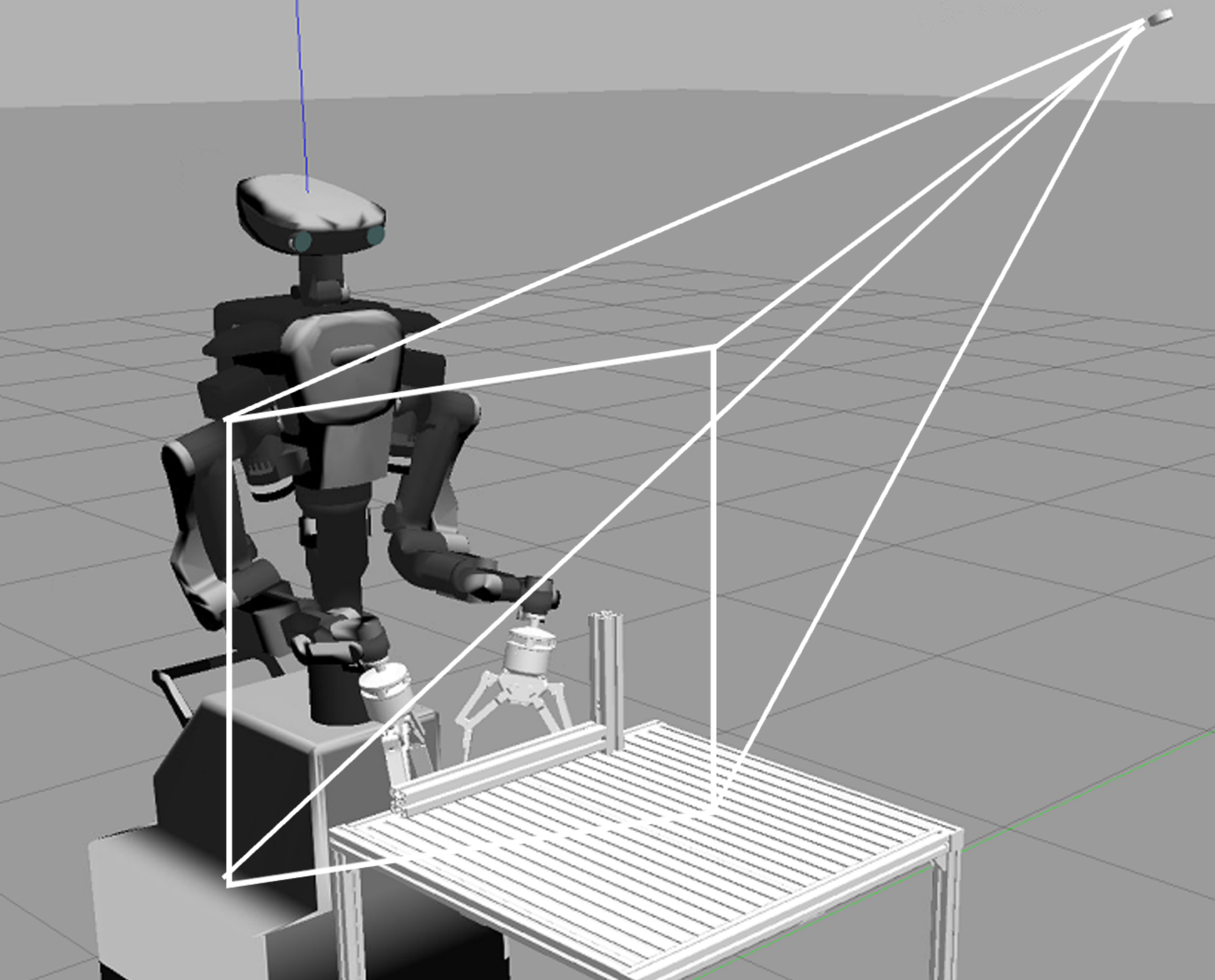}
        \subcaption{Simulation setup}
        \figlab{sim_env}
    \end{minipage}
    \hfill
    \begin{minipage}[tb]{0.39\linewidth}
        \centering
        \includegraphics[width=\linewidth]{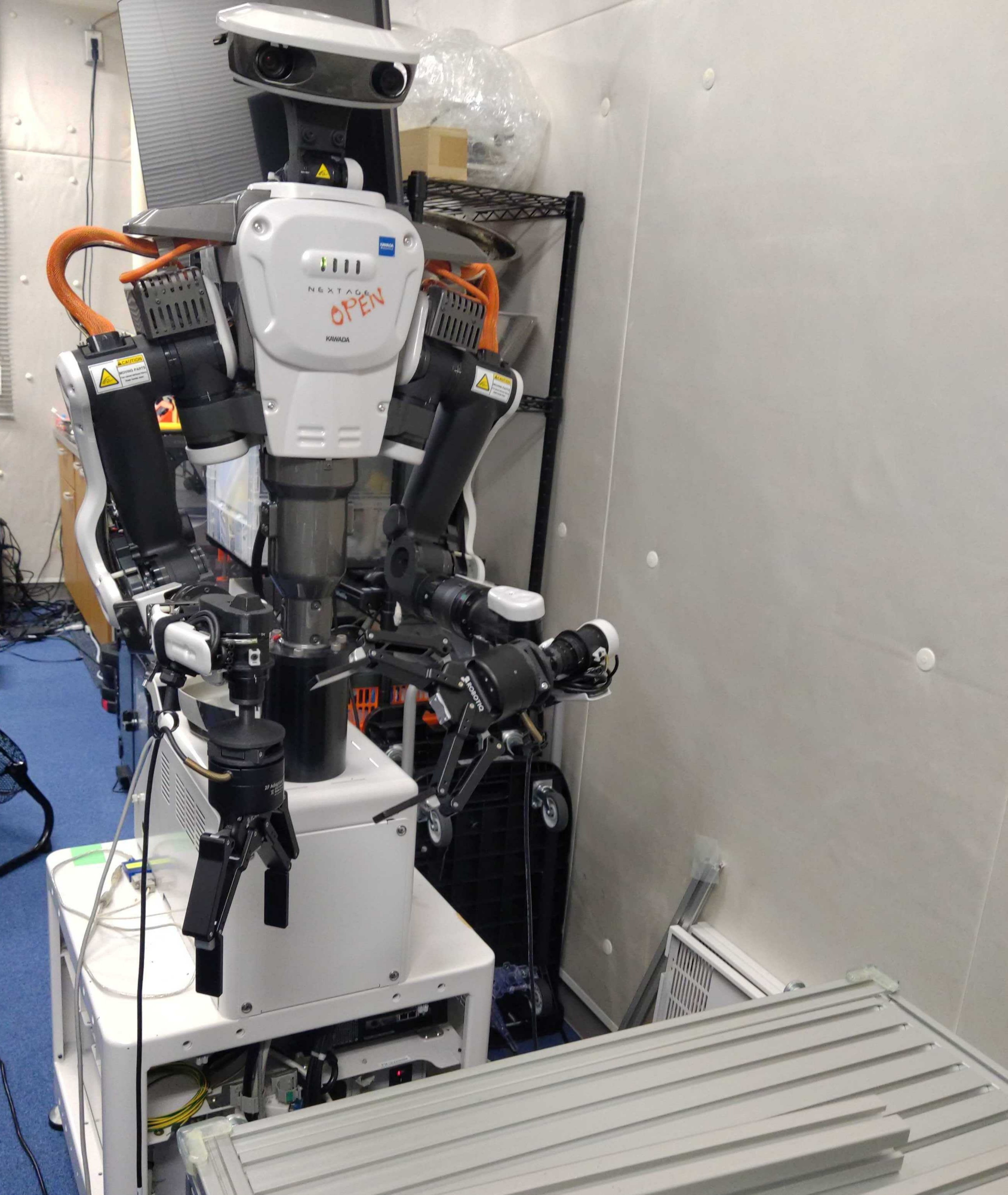}
        \subcaption{Real-world setup}
        \figlab{real_env}
    \end{minipage}
    \caption{Experimental platform used in this study. (a) shows the simulation environment built with Gazebo and ROS, and (b) shows the real-world setup with the Nextage robot.}
    \figlab{setup}
\end{figure*}
The experimental platform is illustrated in \figref{setup}.
We utilized the KAWADA NEXTAGE upper body humanoid robot, which features six degrees of freedom in each arm and one degree of freedom in the waist. A Robotiq 2F-140 two-finger gripper is mounted on the end effector of each arm.
The workspace, constructed on an aluminum frame, contains two aluminum frames to be assembled, a shelf for connection components, and a tool stand for an electric screwdriver.

\figref{setup}~(a) shows the simulation environment, which is implemented using Gazebo, an open-source 3D robot simulator, together with the ROS framework.
\figref{setup}~(b) depicts the real-world setup, which replicates the simulation environment. In the real-world setup, a Logitech C980GR is additionally used to capture images, and audio input is acquired via a built-in PC microphone and converted to text using the Whisper model~\cite{radford2023robust}.

The proposed system integrates three key components: OpenAI o4-mini for high-level reasoning and instruction understanding, GraspGen~\cite{murali2025graspgen} for generating robust 6-DoF grasp candidates, and the RRT-Connect planner implemented in MoveIt~\cite{kuffner2000rrtconnect,chitta2012moveit} for collision-free trajectory planning and trajectory execution.
\hl{For the object-fixation task, GraspGen initially generated 100 6-DoF grasp candidates for each candidate-generation run. We discarded candidates that would collide with the aluminum-frame workbench or surrounding task objects. For each remaining candidate, we used the RRT-Connect planner implemented in MoveIt to plan a trajectory from the robot's initial pose and retained only candidates for which at least one collision-free trajectory was found. After this filtering, 15.1 feasible candidates remained on average per run (SD = 3.3; range: 10--20), and the resulting candidates were used as the \textit{Action Target} candidate set in the experiments.}

We employed the pre-trained o4-mini model in a zero-shot setting without task-specific fine-tuning. For the inference configuration, we set the \texttt{reasoning\_effort} parameter to ``medium'' and the model supports up to 100,000 output tokens. As the temperature parameter is not explicitly controllable in the reasoning-enabled mode, no additional tuning was applied. The input and output formats were structured using JSON to ensure reliable parsing. For the ablation study, we additionally utilized GPT-5.4-mini under two configurations: (1) \texttt{reasoning\_effort} = ``medium'' with 128,000 max output tokens, and (2) \texttt{reasoning\_effort} = ``none'' with \texttt{temperature} = 0.1 and 128,000 max output tokens.
\hl{To improve reproducibility, we used fixed proprietary model snapshots: o4-mini dated April 16, 2025, and GPT-5.4-mini dated March 17, 2026. Both GPT-5.4-mini configurations used the same snapshot. All prompts and inference parameters were fixed across trials.}

\subsection{Tasks and Prompts} \seclab{tasks_prompts}
\begin{figure}[t]
\centering
\small

\begin{tcolorbox}[
  colback=white,
  colframe=black,
  boxrule=0.8pt,
  arc=0pt,
  left=2pt,right=2pt,top=2pt,bottom=2pt
]
\textbf{``\textit{Action Target}'' Selection Model (Object Fixation)}

\begin{tcolorbox}[
  colback=white,
  colframe=black,
  boxrule=0.6pt,
  arc=0pt,
  left=5pt,right=5pt,top=1pt,bottom=1pt
]
\textbf{Task:}\\
You are an AI assistant for assembly tasks.
Given the current object, select a new grasp position candidate according to the human instruction.
If feedback is provided, revise the selection by taking the feedback into account.
\end{tcolorbox}

\vspace{-6pt}

\begin{tcolorbox}[
  colback=white,
  colframe=black,
  boxrule=0.6pt,
  arc=0pt,
  left=5pt,right=5pt,top=1pt,bottom=1pt
  ]
\textbf{Background information:}\\
The following coordinate system definitions take priority over common natural language interpretations.
You must strictly follow these sign conventions and must not interpret them inversely.\\
- Front = -z, Back = +z, Left = +x, Right = -x, Up = +y, Down = -y\\
The displacement is at least 10 percent of the total travel range\\
If it exists, compare it with current grasp
\end{tcolorbox}
\vspace{-6pt}

\begin{tcolorbox}[
  colback=white,
  colframe=black,
  boxrule=0.6pt,
  arc=0pt,
  left=5pt,right=5pt,top=1pt,bottom=1pt
]
\textbf{Input description:}\\
- Instruction: Human language instruction\\
- First object: First grasp\\
- Current object: Current grasp\\
- Action Target candidate: List of grasp candidates[(position[x, y, z], orientation[qx, qy, qz, qw])]\\
- Environment image of the workspace\\
- Previous action history: Prior instructions, executed Action Targets,
and their outcomes (may not exist)\\
- Feedback: (may not exist)
\end{tcolorbox}

\begin{tcolorbox}[
  colback=white,
  colframe=black,
  boxrule=0.6pt,
  arc=0pt,
  left=5pt,right=5pt,top=1pt,bottom=1pt
]
\textbf{Chain of thought:}\\
First, determine the direction and distance of movement based on the instructions, and then select the appropriate \textit{Action Target} from the list of candidates.
\end{tcolorbox}

\vspace{-6pt}

\begin{tcolorbox}[
  colback=white,
  colframe=black,
  boxrule=0.6pt,
  arc=0pt,
  left=5pt,right=5pt,top=1pt,bottom=1pt
]
\textbf{Output format:}\\
New grasp 1\\
New grasp 2\\
New grasp 3\\
Reason: ‘Explanation of selection rationale’
\end{tcolorbox}

\end{tcolorbox}

\caption{Example prompt structure for the Grasp Position Selection Model.}
\figlab{grasp_prompt}
\end{figure}

\begin{figure}[t]
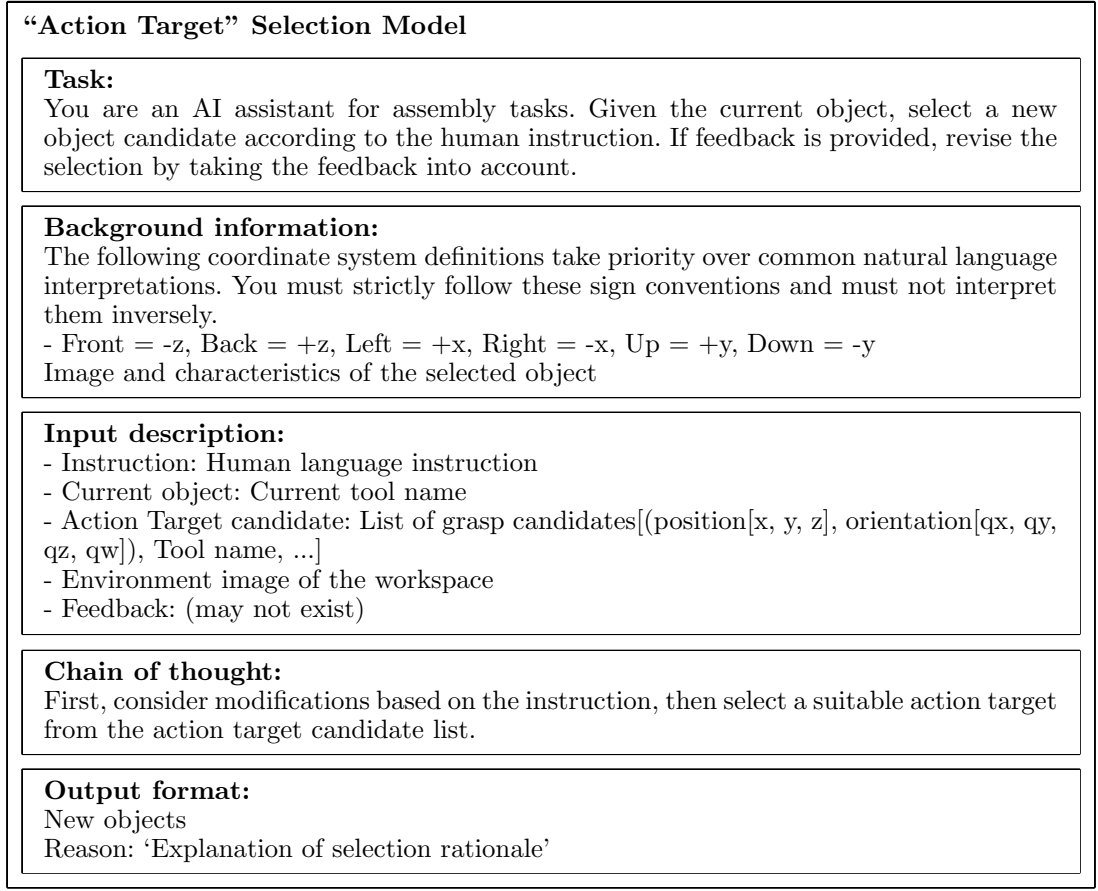

\centering
\small

\begin{tcolorbox}[
  colback=white,
  colframe=black,
  boxrule=0.8pt,
  arc=0pt,
  left=2pt,right=2pt,top=2pt,bottom=2pt
]
\textbf{``\textit{Action Target}'' Selection Model (Tool Preparation)}

\begin{tcolorbox}[
  colback=white,
  colframe=black,
  boxrule=0.6pt,
  arc=0pt,
  left=5pt,right=5pt,top=1pt,bottom=1pt
]
\textbf{Task:}\\
You are an AI assistant for assembly tasks.
Given the current object, select a new tool candidate according to the human instruction.
If feedback is provided, revise the selection by taking the feedback into account.
\end{tcolorbox}

\vspace{-6pt}

\begin{tcolorbox}[
  colback=white,
  colframe=black,
  boxrule=0.6pt,
  arc=0pt,
  left=5pt,right=5pt,top=1pt,bottom=1pt
  ]
\textbf{Background information:}\\
Images and characteristics of available tools
\end{tcolorbox}
\vspace{-6pt}

\begin{tcolorbox}[
  colback=white,
  colframe=black,
  boxrule=0.6pt,
  arc=0pt,
  left=5pt,right=5pt,top=1pt,bottom=1pt
]
\textbf{Input description:}\\
- Instruction: Human language instruction\\
- Current object: Current tool\\
- Action Target candidate: List of tool candidates[toolname]\\
- Environment image of the workspace\\
- Previous action history: Prior instructions, executed Action Targets,
and their outcomes (may not exist)\\
- Feedback: (may not exist)
\end{tcolorbox}

\begin{tcolorbox}[
  colback=white,
  colframe=black,
  boxrule=0.6pt,
  arc=0pt,
  left=5pt,right=5pt,top=1pt,bottom=1pt
]
\textbf{Chain of thought:}\\
None
\end{tcolorbox}

\vspace{-6pt}

\begin{tcolorbox}[
  colback=white,
  colframe=black,
  boxrule=0.6pt,
  arc=0pt,
  left=5pt,right=5pt,top=1pt,bottom=1pt
]
\textbf{Output format:}\\
New tool\\
Reason: ‘Explanation of selection rationale’
\end{tcolorbox}

\end{tcolorbox}

\caption{Example prompt structure for the Tool Selection Model.}
\figlab{tool_prompt}
\end{figure}

\begin{figure}[t]
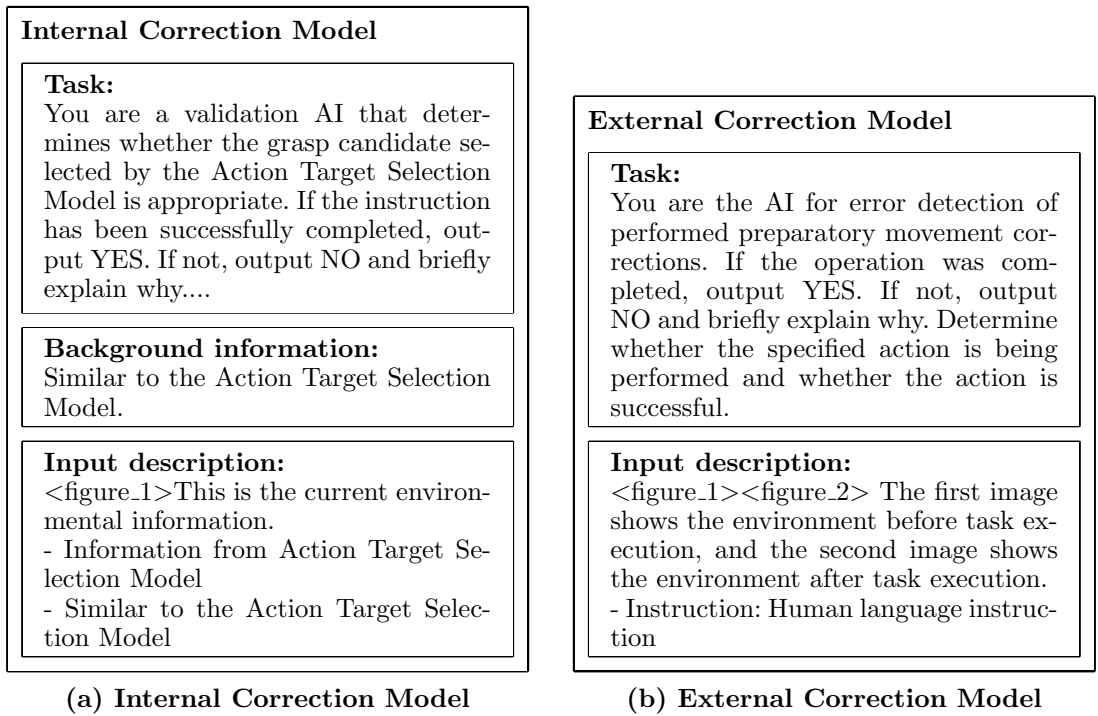

\centering
\small

\begin{minipage}[t]{0.48\linewidth}
\centering

\begin{tcolorbox}[
  colback=white,
  colframe=black,
  boxrule=0.8pt,
  arc=0pt,
  left=2pt,right=2pt,top=2pt,bottom=2pt
]
\textbf{Internal Correction Model}

\begin{tcolorbox}[colback=white,colframe=black,boxrule=0.6pt,arc=0pt,left=5pt,right=5pt,top=1pt,bottom=1pt]
\textbf{Task:}\\
You are a validation AI that determines whether the grasp candidate selected by the Action Target Selection Model is appropriate. If the instruction has been successfully completed, output YES. If not, output NO and briefly explain why....
\end{tcolorbox}

\vspace{-6pt}

\begin{tcolorbox}[colback=white,colframe=black,boxrule=0.6pt,arc=0pt,left=5pt,right=5pt,top=1pt,bottom=1pt]
\textbf{Background information:}\\
Similar to the Action Target Selection Model.
\end{tcolorbox}

\vspace{-6pt}

\begin{tcolorbox}[colback=white,colframe=black,boxrule=0.6pt,arc=0pt,left=5pt,right=5pt,top=1pt,bottom=1pt]
\textbf{Input description:}\\
$<$figure\_1$>$This is the current environmental information.\\
- Information from Action Target Selection Model\\
- Similar to the Action Target Selection Model
\end{tcolorbox}

\end{tcolorbox}

\textbf{(a) Internal Correction Model}
\end{minipage}
\hfill
\begin{minipage}[t]{0.48\linewidth}
\centering

\begin{tcolorbox}[
  colback=white,
  colframe=black,
  boxrule=0.8pt,
  arc=0pt,
  left=2pt,right=2pt,top=2pt,bottom=2pt
]
\textbf{External Correction Model}

\begin{tcolorbox}[colback=white,colframe=black,boxrule=0.6pt,arc=0pt,left=5pt,right=5pt,top=1pt,bottom=1pt]
\textbf{Task:}\\
You are the AI for error detection of performed preparatory movement corrections.
If the operation was completed, output YES. If not, output NO and briefly explain why.
Determine whether the specified action is being performed and whether the action is successful.
\end{tcolorbox}

\vspace{-6pt}

\begin{tcolorbox}[colback=white,colframe=black,boxrule=0.6pt,arc=0pt,left=5pt,right=5pt,top=1pt,bottom=1pt]
\textbf{Input description:}\\
$<$figure\_1$>$$<$figure\_2$>$ The first image shows the environment before task execution,
and the second image shows the environment after task execution.\\
- Instruction: Human language instruction
\end{tcolorbox}

\end{tcolorbox}

\textbf{(b) External Correction Model}
\end{minipage}

\caption{Example prompt structures for Internal and External Correction models.}
\figlab{prompt_models}
\end{figure}
In the object fixation task, the robot assists a human in the L-shaped assembly of two aluminum frames. While the human worker attaches a bracket to connect the frames, the robot stabilizes one of the frames to ensure a steady work environment. In the real-world setup, as the frame may tip over during the robot's approach or pose adjustment, the worker provides temporary support until the grasp is fully secure. During this process, the human provides instructions such as ``Move a little more to the left'' or ``Raise the left a bit higher''. Success is defined as the robot correctly modifying its grasp pose based on these spatial commands without causing collisions.

In the tool preparation task, the robot retrieves an electric screwdriver from a stand and hands it over to the worker. The stand holds six screwdrivers, each equipped with a different-sized hex bit for specific bolts. The task begins with an initial command such as ``Take a hex driver''. We also evaluated the system's dynamic replanning capability by providing corrective instructions after the initial selection, such as ``Take a bigger one''. Success is defined as the robot's ability to identify, retrieve, and if necessary, switch to the correct tool according to the linguistic input.

To clarify the VLM's input and output structures, we detail the prompt designs for each module. As shown in \figref{grasp_prompt} and \figref{tool_prompt}, the prompts for the \textit{Action Target} Selection Models consist of the task description, background information, input description, a chain-of-thought directive, and the expected output format. The object fixation task explicitly defines the correspondence between the robot and human coordinate systems to guide spatial reasoning, whereas the tool preparation task provides the characteristics of available tools. Additionally, as shown in \figref{prompt_models}, the Internal and External Correction models receive task-specific verification criteria along with environmental images, with the Internal Correction Model also leveraging background information for logical consistency checks.

\subsection{Results} \seclab{results}
\subsubsection{Quantitative Analysis}
\begin{table}[tb]
    \centering
    \caption{Success rates in simulation experiments.}
    \tablab{sim_results}
    \begin{tabular}{llc}
    \toprule
    \textbf{Task} & \textbf{Instruction Category} & \textbf{Success Rate} \\
    \midrule
    Object Fixation & Position adjustment (Left / Right) & 20/20 (100\%) \\
    Tool Preparation & Initial selection & 10/10 (100\%) \\
    Tool Preparation & Corrective selection (Replanning) & 9/10 (90.0\%) \\
    \bottomrule
    \end{tabular}
\end{table}
\begin{table}[tb]
    \centering
    \caption{Success rates in real-world experiments.}
    \tablab{real_results}
    \begin{tabular}{llc}
    \toprule
    \textbf{Task} & \textbf{Instruction Category} & \textbf{Success Rate} \\
    \midrule
    Object Fixation & Position adjustment & 10/15 (66.7\%) \\
    Tool Preparation & Initial selection & 8/8 (100\%) \\
    Tool Preparation & Corrective selection (Replanning) & 6/8 (75.0\%) \\
    \bottomrule
    \end{tabular}
\end{table}
\tabref{sim_results} and \tabref{real_results} summarize the success rates in simulation and real-world environments, respectively.

In the simulation experiments, the proposed method achieved a 100\% success rate (20/20 trials) for position adjustment in the object fixation task. For the tool preparation task, the system demonstrated perfect performance in the initial selection (10/10 trials) and achieved a 90\% success rate (9/10 trials) in corrective selection involving replanning. A minor failure occurred in corrective selection due to visual occlusion, where the robot's own arm occasionally obstructed the camera's view of the remaining tools, leading to a misinterpretation by the External Correction Model.

In the real-world experiments, the object-fixation task resulted in a 66.7\% success rate (10/15 trials). \hl{Inspection of the failed trials suggested that, when the robot arm and gripper approached the target aluminum frame, they partially occluded the frame in the camera view, sometimes compounded by lighting variations. This made the intended change in the relative pose between the gripper and the frame difficult for the External Correction Model to verify from the before-and-after images, leading to false-negative judgments and unnecessary replanning.} For the tool preparation task, the system achieved a 100\% success rate (8/8 trials) in the initial selection. In the corrective selection phase, such as responding to instructions to change the tool size, the success rate was 75\% (6/8 trials). These failures were primarily attributed to lighting conditions where the white handle of the driver became indistinguishable from the background, hindering the External Correction Model's ability to verify the state change. \hl{Because these causes were inferred from observed failures rather than controlled comparisons, the individual effects of lighting and occlusion on visual verification should be systematically evaluated in future work.}

\begin{table}[tb]
    \centering
    \caption{Performance comparison across different VLM configurations (100 action steps).}
    \tablab{vlm_comparison}
    \resizebox{\columnwidth}{!}{%
    \begin{tabular}{l|cc|cc|cc}
    \toprule
    \multirow{2}{*}{\textbf{Correction Modules}} & \multicolumn{2}{c|}{\textbf{o4-mini}} & \multicolumn{2}{c|}{\textbf{5.4-mini (\texttt{medium})}} & \multicolumn{2}{c}{\textbf{5.4-mini (\texttt{none})}} \\
    & Task Succ. & Sel. Succ. & Task Succ. & Sel. Succ. & Task Succ. & Sel. Succ. \\
    \midrule
    w/o Both (Baseline) & 94\% & 86\% & 99\% & 97\% & 87\% & 85\% \\
    w/o Internal & 61\% & - & \textbf{100\%} & - & 83\% & - \\
    w/o External & \textbf{100\%} & \textbf{100\%} & \textbf{100\%} & \textbf{100\%} & 89\% & \textbf{87\%} \\
    \textbf{Full (Internal + External)} & 72\% & \textbf{100\%} & \textbf{100\%} & \textbf{100\%} & \textbf{94\%} & \textbf{87\%} \\
    \midrule
    Avg. Inference Time (Selection) & \multicolumn{2}{c|}{17.7 s} & \multicolumn{2}{c|}{19.5 s} & \multicolumn{2}{c}{3.59 s} \\
    \bottomrule
    \end{tabular}%
    }
\end{table}

To further evaluate the robustness against continuous interaction and justify our choice of the primary VLM, we conducted an ablation study comparing our primary model (o4-mini) with GPT-5.4-mini under two reasoning configurations (\texttt{medium} and \texttt{none}). The evaluation comprised 100 action steps in total, organized as
20 trials of five consecutive spatial adjustments. We evaluated the models using two metrics: \textit{Task Success Rate} (successful execution of the correct motion) and \textit{Selection Success Rate}. The latter strictly requires all three generated grasp candidates to be geometrically and logically valid. This strict metric is important for real-world applications, where fallback candidates may be needed if the top-ranked candidate becomes infeasible due to updated scene conditions or execution constraints.

\hl{As shown in} \tabref{vlm_comparison}\hl{, the results demonstrate distinct characteristics depending on the model's reasoning capability. The reasoning-enabled configurations (o4-mini and 5.4-mini medium) achieved high Task Success Rates in the no-correction baseline, but still occasionally generated invalid or hallucinated candidates. For these models, the Internal Correction model successfully filtered out all invalid candidates, achieving a 100\% Selection Success Rate. For o4-mini, however, the External Correction Model sometimes judged successfully executed motions as failures due to visual ambiguity in the simulated before-and-after images. This visual false-negative issue reduced the Task Success Rate of the full configuration to 72\%, compared with 94\% without correction and 100\% with the Internal Correction only.}

\hl{Conversely, the low-latency model (5.4-mini none) showed exceptionally fast inference (average 3.59 s for selection compared to 17.7 s for o4-mini and 19.5 s for the medium configuration) but demonstrated lower baseline stability, frequently proposing opposite directions. In this configuration, the full dual-correction framework integrating both internal and external modules was effective, improving the Task Success Rate from 87\% to 94\% and achieving the highest overall success rate for this specific model.}

These comparative results highlight a clear trade-off between reasoning stability and operational latency. Reasoning models like o4-mini provide robust baseline accuracy suitable for complex tasks, whereas low-latency non-reasoning models offer rapid responses but heavily depend on the dual-correction mechanism to ensure safety. Therefore, the optimal VLM configuration should be selectively adopted based on whether the specific collaborative scenario prioritizes immediate responsiveness or strict reasoning stability.

Finally, we validated the intrinsic performance of the correction models using adversarial inputs, such as instructing the robot to move left when the visual state showed no movement. Both the Internal and External Correction models achieved a 100\% detection rate (10/10 trials each) in these isolated tests, confirming their fundamental capability to detect logical and physical inconsistencies when environmental noise is controlled.

\subsubsection{Qualitative Analysis}
\begin{figure*}[tb]
    \centering
    \begin{minipage}[tb]{0.48\linewidth}
        \centering
        \includegraphics[width=\linewidth]{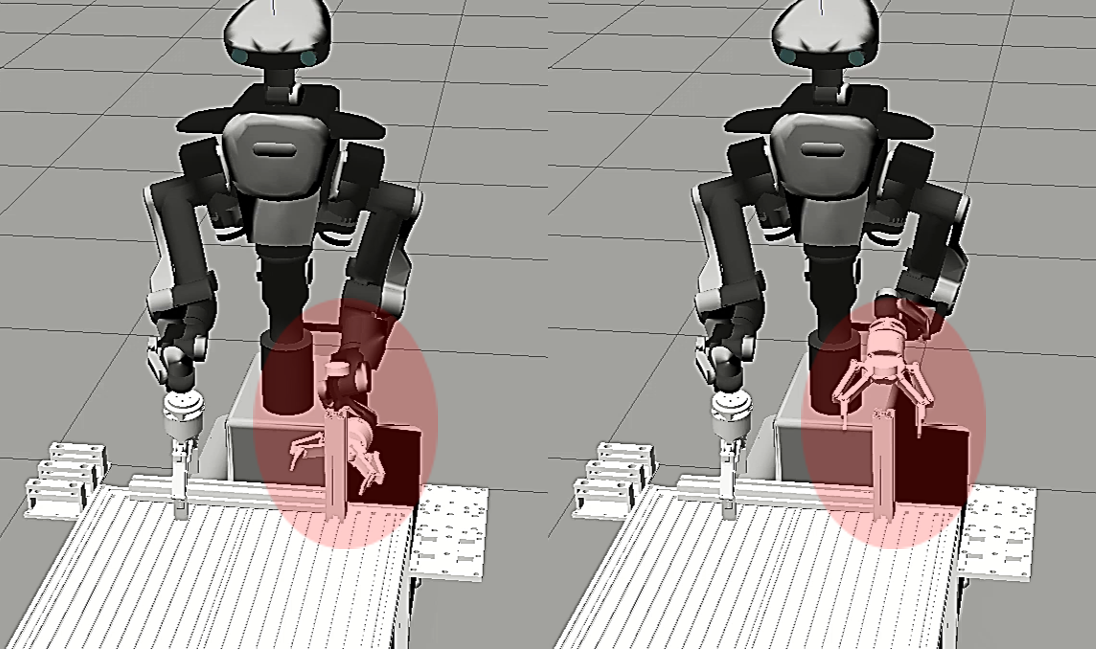}
        \subcaption{Left arm adjustment: ``Raise the left a bit higher''}
        \figlab{sim_fix_left}
    \end{minipage}
    \hfill
    \begin{minipage}[tb]{0.48\linewidth}
        \centering
        \includegraphics[width=\linewidth]{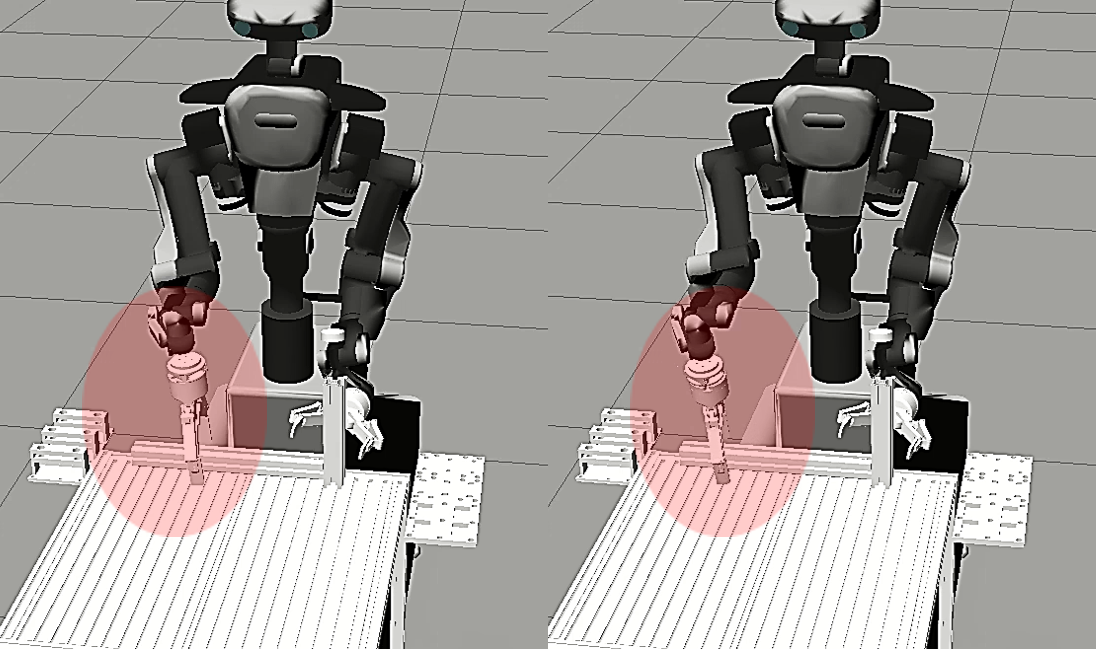}
        \subcaption{Right arm adjustment: ``Move a little more to the left''}
        \figlab{sim_fix_right}
    \end{minipage}
    \caption{A successful object fixation task in simulation. The robot adjusts its grasp pose to stabilize an aluminum frame for subsequent manual assembly. Red highlights indicate the target end-effector movements corresponding to the linguistic instructions.}
    \figlab{sim_fix_results}
\end{figure*}
\begin{figure*}[tb]
    \centering
    \begin{minipage}[tb]{\linewidth}
        \centering
        \includegraphics[width=\linewidth]{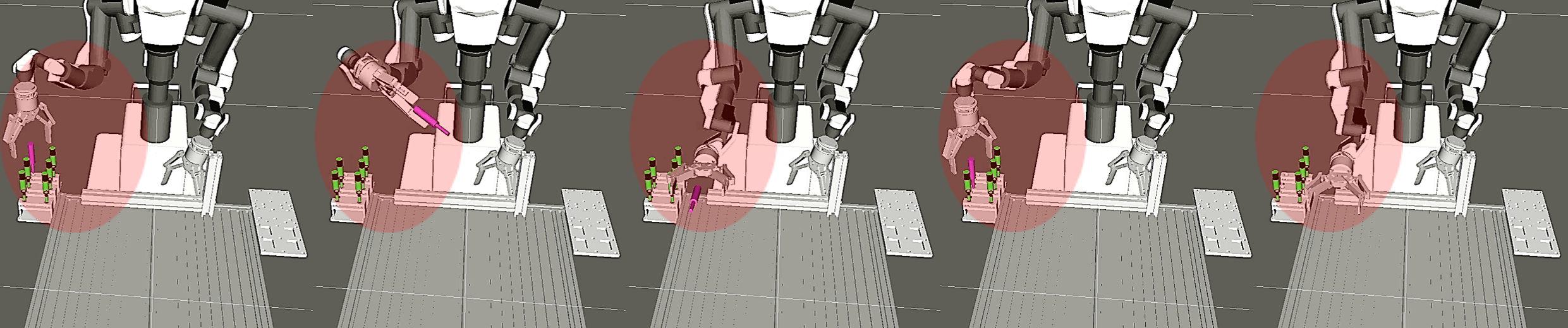}
        \subcaption{Initial selection phase}
        \figlab{sim_tool_1}
    \end{minipage}
    \vspace{1em}
    \begin{minipage}[tb]{\linewidth}
        \centering
        \includegraphics[width=\linewidth]{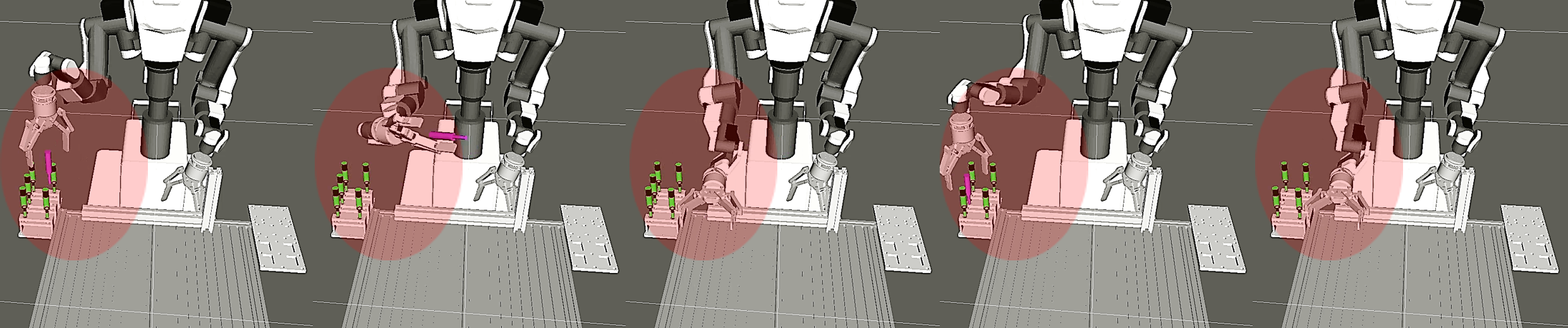}
        \subcaption{Replanning phase with corrective instruction}
        \figlab{sim_tool_2}
    \end{minipage}
    \caption{Successful execution of the tool preparation task demonstrating dynamic replanning in simulation. The robot selects and hands over a specific hex driver from a tool stand containing six candidates with different bit sizes. Red highlights track the sequence of grasping, corrective instruction receipt (\eg, switching to a larger size), and successful replanning.}
    \figlab{sim_tool_results}
\end{figure*}
\begin{figure*}[tb]
    \centering
    \includegraphics[width=\linewidth]{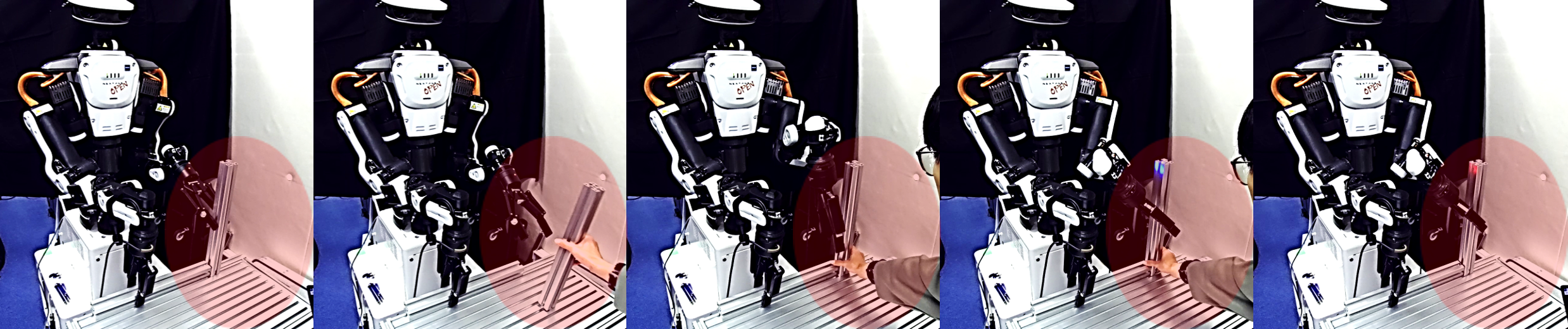}
    \caption{Real-world execution of the object fixation task, corresponding to the scenario in~\figref{sim_fix_results}~(a). The robot adjusts the grasping pose of an aluminum frame while the human worker provides temporary support to prevent it from tipping during the motion. Red highlights indicate the coordinated adjustment of the end-effector and the frame.}
    \figlab{real_fix_result}
\end{figure*}
\begin{figure*}[tb]
    \centering
    \includegraphics[width=\linewidth]{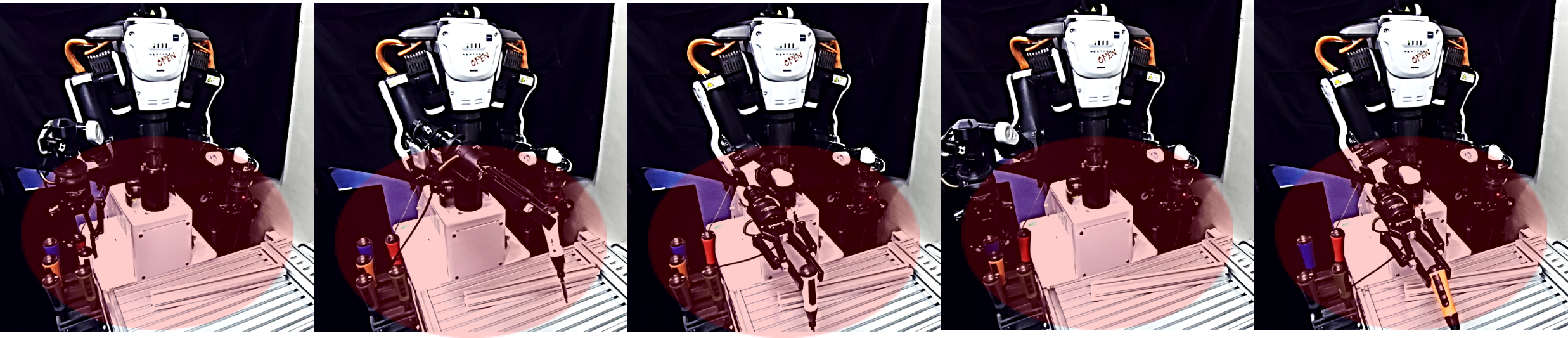}
    \caption{Real-world tool preparation task, corresponding to the replanning sequence in~\figref{sim_tool_results}~(b). The robot identifies the correct screwdriver among six candidates and hands it to the human worker. Red highlights denote the target tool selection and the subsequent handover action.}
    \figlab{real_tool_result}
\end{figure*}
\figref{sim_fix_results} demonstrates the qualitative results of the fixation task in simulation. As shown in \figref{sim_fix_results}~(a), upon receiving the instruction ``Raise the left a bit higher,'' the VLM selected an \textit{Action Target} corresponding to an upward offset, and the robot correctly adjusted the left arm's grasping position. Similarly, \figref{sim_fix_results}~(b) shows the robot shifting the right arm in response to ``Move a little more to the left.'' These cases confirm that the VLM correctly interprets spatial directions relative to the robot's coordinate frame and maps them to the appropriate action candidates.

\figref{sim_tool_results} illustrates the dynamic replanning capability in the tool preparation task. Initially, the robot grasped a standard hex driver as instructed (\figref{sim_tool_results}~(a)). However, when provided with the ambiguous corrective instruction ``Take a bigger one,'' the system successfully inferred the user's intent by comparing the remaining candidates and switched to a driver equipped with a larger hex bit (\figref{sim_tool_results}~(b)). This transition demonstrates that retaining previous action history in the persistent reasoning context $\ell_{\text{current}}$ enables the system to interpret comparative linguistic concepts.

\figref{real_fix_result} presents the result of the fixation task in the real-world environment. As shown in the figure, the robot successfully performed physical adjustments, demonstrating the transferability of the generated grasps. However, we also observed false-negative cases during the experiments. In these cases, lighting variations and partial occlusion by the robot arm made the intended change in the relative pose between the gripper and the target frame difficult to verify from the before-and-after images. As a result, the External Correction Model incorrectly triggered replanning despite the robot having reached the intended pose.

Finally, \figref{real_tool_result} shows the execution of the tool preparation task in the real world. The robot successfully identified the target tool from the stand. During the replanning phase, the system executed a sequence of releasing the current tool and returning to the selection phase to grasp the new target, validating the feasibility in a real-world setting.

\section{Discussion} \seclab{discussion}

\subsection{Robustness of the Dual-Correction Mechanism}
Experimental results characterize the robustness and limitations of the proposed framework. The ablation study on consecutive instructions indicates that while single-turn tasks may succeed without correction, multi-turn interactions depend heavily on it, particularly when using low-latency models. Across the 20 five-step object-fixation trials, the task success rate for the non-reasoning VLM configuration decreased from 94\% to 87\% upon removing the dual-correction modules. Furthermore, for reasoning models like o4-mini, the Internal Correction Model proved essential for filtering out hallucinated candidates, recovering the selection success rate from 86\% to 100\%. This is consistent with Huang~\etal~\cite{huang2022inner}, who showed that closed-loop environmental feedback is essential for grounding LLM reasoning over long horizons. \hl{Our results confirm that minor errors, negligible in a single step, can accumulate over time; thus, correction mechanisms are useful for maintaining consistency, particularly in low-latency configurations, as noted in ReplanVLM}~\cite{mei2024replanvlm}\hl{. However, the o4-mini result also shows that External Correction can reduce task success when the intended state change is not clearly observable in the before-and-after images.}

\hl{It should be noted that} \tabref{vlm_comparison} \hl{reports controlled ablation studies rather than a direct comparison with independently implemented competing methods. This design was chosen to isolate the effects of the Internal and External Correction modules under identical VLM inputs, \textit{Action Target} candidates, grasp generation, and motion-planning conditions. Direct comparison with adapted existing methods under the same execution pipeline remains future work.}

The isolated validation tests suggest that the correction models can detect logical and physical inconsistencies when visual noise is controlled. The remaining failures were mainly associated with two factors: visual false negatives in the External Correction Model under lighting changes or occlusion, and invalid output formats from the \textit{Action Target} Selection Model when internal verification was removed. We leave systematic quantification of these visual factors for future work. To avoid unsafe or indefinite execution, the framework terminates the task after three rejections by the Internal Correction Model or after three consecutive physical failures identified by the External Correction Model.

These results highlight visual-state verification as a key limitation of the current framework. To bridge this gap, more robust perception and state-verification modules are needed. To mitigate these issues, the \textit{Action Target} Selection Model's accuracy could be enhanced through fine-tuning on assembly-specific datasets. Additionally, the current verification process updates the environment image after each failure, which can cause the model to lose track of the initial state. Retaining the initial state image alongside the current one may improve the recognition of relative changes. Furthermore, pre-processing steps, such as labeling the end-effector and target objects in the input images, would enhance object recognition and overall success rates.

\subsection{Operational Latency}
In our experimental setup, the total latency from instruction input to execution, including the Selection and Internal Correction models, was approximately 30 seconds. The Selection Model required an average of 17.7 seconds, while the Internal Correction Model required approximately 13 seconds, resulting in a total latency of approximately 30 seconds. This similarity in processing time is attributed to the comparable lengths of their respective prompts, input images, and candidate lists. The External Correction Model completed its verification in under 10 seconds, likely due to its simpler input structure consisting only of two images and the linguistic instruction. 

While these processing times are sufficient for the current assembly tasks, as demonstrated in the comparative analysis (see \secref{results}), the operational latency is highly dependent on the chosen VLM architecture. Transitioning to non-reasoning configurations (\eg, 5.4-mini \texttt{none}) can drastically reduce the inference time per correction loop to a few seconds. However, because such low-latency models exhibit lower baseline stability, balancing this trade-off through our dual-correction mechanism and optimal model selection will be critical for time-critical interactions in dynamic environments.

\subsection{Task Success Criteria and Human Factors} \subseclab{human_factors}
A fundamental objective of this framework is to generate assistive motions from ambiguous human instructions (e.g., ``move a little more to the left''). Because such instructions inherently lack explicit numerical ground truth, defining task success strictly through quantitative positional thresholds is challenging. In human-robot collaboration, task success should ideally be evaluated based on the intuitive satisfaction and intent fulfillment of the human co-worker. As a future prospect, we aim to establish a user-customized assistive motion generation method. Inferring personalized preferences from user interactions, an approach successfully demonstrated by TidyBot~\cite{wu2023tidybot}, represents a promising solution. Learning individual worker preferences would not only improve collaborative efficiency but also facilitate the establishment of clearer evaluation criteria. Specifically, by mapping subjective expressions to user-specific numerical ranges, the system could define a quantifiable baseline. This would allow task success to be objectively measured against a user's personalized intent rather than an elusive universal standard.

Furthermore, the system's passive nature remains a limitation related to human factors. This reactive behavior can impose a high cognitive load on the human worker. By integrating the VLM with long-context memory or task planners, such as KnowNo~\cite{ren2023robots}, the robot could infer assembly sequences and anticipate the next \textit{Action Target}. Proactive behaviors, such as preparing a tool before an explicit request, would facilitate a more natural human--robot partnership. Utilizing this assistive robot as a collaborative mechanical interface necessitates a rigorous evaluation of the human workload. In our future research, we plan to quantitatively measure the impact of this passive interaction versus proactive assistance on human operational efficiency and cognitive load, utilizing standardized assessment tools such as the NASA Task Load Index (NASA-TLX).

\subsection{Future Extension: Force Adaptation and Universality} \subseclab{force_universality}

\hl{This subsection presents a preliminary future-work exploration and is not part of the core quantitative evaluation of the proposed position-controlled \textit{Action Target} replanning framework.}

Although the proposed framework ensures geometric accuracy, the current reliance on position control is insufficient for tasks requiring physical compliance, such as component insertion. To address this, the robot must be able to interpret force-related nuances like ``hold it tightly.'' Recent studies suggest that language-conditioned models can connect high-level instructions to motion and control: VoxPoser~\cite{huang2023voxposer} uses LLMs and VLMs to compose 3D value maps for motion planning, whereas Language-to-Rewards~\cite{yu2023language} uses LLMs to generate reward functions from language instructions. Extending such language-conditioned reasoning to the force domain would enable autonomous tuning of impedance parameters---prioritizing high stiffness for assembly feasibility or high compliance for safety.

To explore the feasibility of such language-guided force adaptation, we conducted a preliminary experiment mapping ambiguous instructions to grasp-related parameters (opening width and grasping force). Given a system prompt defining parameter ranges, the VLM intuitively adjusted these values from a baseline state: it decreased the force for ``hold it lightly'' and increased both width and force for ``hold it firmly.'' Furthermore, it successfully reflected the nuances of adverbial modifiers, applying proportionately smaller changes for ``loosen it slightly'' or ``tighten it a little.'' While these results provide promising initial evidence, accurately mapping nuanced linguistic instructions to precise, multi-dimensional impedance control parameters remains a key challenge for future studies.

Regarding universality, our visual reasoning and standard trajectory planning can be readily applied to other robot platforms, such as single-arm manipulators, provided the end-effector maintains a consistent mechanical configuration. However, for assistive tasks requiring collaborative transportation or precise force control, changing the robot arm significantly alters the underlying kinematics and dynamics. Therefore, verifying the framework's universality across diverse platforms under varying dynamic constraints constitutes another essential direction for future work.

\section{Conclusion}
This study presented a robust replanning framework for human--robot collaborative assembly by integrating Vision--Language Models (VLMs) with a dual-correction mechanism. By defining assistive tasks through the abstraction of \textit{Action Targets}, we developed a unified pipeline capable of handling diverse collaborative scenarios, including object fixation and tool preparation. The integration of Internal and External Correction models formed a closed-loop system that autonomously detects logical hallucinations and physical execution failures, enabling reliable error recovery through iterative reasoning. 

\hl{Experimental results in both simulation and real-world environments showed configuration-dependent benefits of the correction modules. Internal Correction improved candidate validity, while External Correction supported post-execution recovery for the low-latency VLM configuration but could reduce task success when visual false negatives occurred.} While the system demonstrated high adaptability to ambiguous linguistic instructions, real-world trials also highlighted the sensitivity of visual verification to lighting and occlusion. \hl{Future work will focus on improving visual verification, incorporating long-term context awareness, and extending the framework toward language-guided force adaptation, which was discussed as a preliminary future-work direction in this study.}

\section*{Data availability statement}
\hl{The code and raw experimental data are not publicly available at this stage because the implementation includes hardware-dependent robot-control and safety-related components. We plan to release them after further robustness and safety validation.}

\section*{Funding}
This work was supported by JST Moonshot R\&D Program Grant Number JPMJMS263E-9.

\section*{Disclosure statement}
No potential conflict of interest was reported by the authors.

\section*{Declaration of generative AI use}
OpenAI o4-mini and GPT-5.4-mini were used as components of the experimental system described in this study. ChatGPT (OpenAI) was used for English-language editing and revision suggestions during manuscript preparation. All AI-assisted text and technical statements were reviewed and verified by the authors, who take full responsibility for the final manuscript.

\section*{Notes on contributors}
\noindent\textit{\textbf{Taichi Kato}} is currently pursuing the Bachelor's degree in the School of Engineering Science, the University of Osaka, Toyonaka, Japan. His current research interests include human--robot collaboration with vision--language models.
\vspace{2mm}

\noindent\textit{\textbf{Takuya Kiyokawa}} received the B.E. degree from the National Institute of Technology, Kumamoto College, Japan, and the M.E. degree and the Ph.D. degree in engineering from the Nara Institute of Science and Technology, Japan, in 2018 and 2021, respectively. From 2021 to 2022, he was with The University of Osaka, Japan, as a Specially-Appointed Assistant Professor, and with the Nara Institute of Science and Technology, as a Specially-Appointed Assistant Professor. From 2023 to 2024, he was a Visiting Researcher with the Institute of Robotics and Mechatronics, German Aerospace Center (DLR), Oberpfaffenhofen, Weßling, Germany and since 2023, he has been with The University of Osaka, as an Assistant Professor. His current research interests include robot manipulation and agile reconfigurable robotic systems.
\vspace{2mm}

\noindent\textit{\textbf{Namiko Saito}} received the B.S., M.S., and Ph.D. degrees in mechanical engineering from Waseda University, Japan, in 2018, 2020, and 2022, respectively. From 2023 to 2024, she was a Research Associate at The University of Edinburgh, affiliated with the Alan Turing Institute, UK. She is currently a Senior Researcher at Microsoft Research Asia – Tokyo, Japan. Her research interests include embodied AI, robot manipulation, multimodal learning, and sensorimotor control.
\vspace{2mm}

\noindent\textit{\textbf{Kensuke Harada}} received the B.Sc., M.Sc., and Ph.D. degrees in mechanical engineering from Kyoto University, Kyoto, Japan, in 1992, 1994, and 1997, respectively. From 1997 to 2002, he was a Research Associate with Hiroshima University, Hiroshima, Japan. Since 2002, he has been with the National Institute of Advanced Industrial Science and Technology (AIST). From 2005 to 2006, he was a Visiting Scholar with the Department of Computer Science, Stanford University, Stanford, CA, USA, and the Leader of the Manipulation Research Group, AIST, from 2013 to 2015. He is currently a Professor with the Graduate School of Engineering Science, The University of Osaka, Toyonaka, Japan. His research interests include mechanics and control of robot manipulators and robot hands, biped locomotion, and motion planning of robotic systems.

\bibliographystyle{tfnlm}
\bibliography{interactnlmsample}

@article{coronado2022evaluating,
  title={Evaluating quality in human-robot interaction: A systematic search and classification of performance and human-centered factors, measures and metrics towards an industry 5.0},
  author={Coronado, Enrique and Kiyokawa, Takuya and Ricardez, Gustavo A Garcia and Ramirez-Alpizar, Ixchel G and Venture, Gentiane and Yamanobe, Natsuki},
  journal={Journal of Manufacturing Systems},
  volume={63},
  pages={392--410},
  year={2022}
}

@article{kotake2025cooperation,
  title={A Cooperation Control Framework Based on Admittance Control and Time-Varying Passive Velocity Field Control for Human--Robot Co-Carrying Tasks},
  author={Kotake, Hiroki and Honji, Sumitaka and Wada, Takahiro and others},
  journal={IEEE Transactions on Automation Science and Engineering},
  volume={22},
  pages={23579--23593},
  year={2025}
}

@article{wang2024assembly,
  title={Assembly Task Allocation for Human-Robot Collaboration Considering Stability and Assembly Complexity},
  author={Wang, Zhenting and Kiyokawa, Takuya and Yamanobe, Natsuki and Wan, Weiwei and Harada, Kensuke},
  journal={IEEE Access},
  year={2024},
  volume={12},
  pages={159821--159832}
}

@article{ji2024foundation,
  title={Foundation models assist in human--robot collaboration assembly},
  author={Ji, Yuchen and Zhang, Zequn and Tang, Dunbing and Zheng, Yi and Liu, Changchun and Zhao, Zhen and Li, Xinghui},
  journal={Scientific Reports},
  volume={14},
  number={1},
  pages={24828},
  year={2024}
}

@inproceedings{lim2024enhancing,
  title={Enhancing human-robot collaborative assembly in manufacturing systems using large language models},
  author={Lim, Jonghan and Patel, Sujani and Evans, Alex and Pimley, John and Li, Yifei and Kovalenko, Ilya},
  booktitle={Proceedings of IEEE International Conference on Automation Science and Engineering (CASE)},
  pages={2581--2587},
  year={2024}
}

@article{liu2024vision,
  title={Vision AI-based human-robot collaborative assembly driven by autonomous robots},
  author={Liu, Sichao and Zhang, Jianjing and Wang, Lihui and Gao, Robert X},
  journal={CIRP annals},
  volume={73},
  number={1},
  pages={13--16},
  year={2024}
}

@article{liu2024enhancing,
  title={Enhancing the llm-based robot manipulation through human-robot collaboration},
  author={Liu, Haokun and Zhu, Yaonan and Kato, Kenji and Tsukahara, Atsushi and Kondo, Izumi and Aoyama, Tadayoshi and Hasegawa, Yasuhisa},
  journal={IEEE Robotics and Automation Letters},
  volume={9},
  Issue={8},
  pages={6904--6911},
  year={2024}
}

@article{mei2024replanvlm,
  title={Replanvlm: Replanning robotic tasks with visual language models},
  author={Mei, Aoran and Zhu, Guo-Niu and Zhang, Huaxiang and Gan, Zhongxue},
  journal={IEEE Robotics and Automation Letters},
  volume={9},
  Issue={11},
  pages={10201--10208},
  year={2024}
}

@inproceedings{black2024pi_0,
  title={$\pi_0 $: A Vision-Language-Action Flow Model for General Robot Control},
  author={Black, Kevin and Brown, Noah and Driess, Danny and Esmail, Adnan and Equi, Michael and Finn, Chelsea and Fusai, Niccolo and Groom, Lachy and Hausman, Karol and Ichter, Brian and others},
  booktitle={Proceedings of Robotics: Science and Systems},
  year={2025}
}

@inproceedings{murali2025graspgen,
  title     = {{GraspGen}: A Diffusion-based Framework for 6-DOF Grasping with On-Generator Training},
  author    = {Murali, Adithyavairavan and Sundaralingam, Balakumar and Chao, Yu-Wei and Yamada, Jun and Yuan, Wentao and Carlson, Mark and Ramos, Fabio and Birchfield, Stan and Fox, Dieter and Eppner, Clemens},
  booktitle = {Proceedings of the IEEE International Conference on Robotics and Automation (ICRA)},
  year      = {2026},
}

@inproceedings{radford2023robust,
  title={Robust speech recognition via large-scale weak supervision},
  author={Radford, Alec and Kim, Jong Wook and Xu, Tao and Brockman, Greg and McLeavey, Christine and Sutskever, Ilya},
  booktitle={Proceedings of International Conference on Machine Learning},
  pages={28492--28518},
  year={2023}
}

@inproceedings{ahn2022do,
  title={Do as {I} can, not as {I} say: Grounding language in robotic affordances},
  author={Ichter, Brian and Brohan, Anthony and Chebotar, Yevgen and Finn, Chelsea and Hausman, Karol and Herzog, Alexander and Ho, Daniel and Ibarz, Julian and Irpan, Alex and Jang, Eric and Julian, Ryan and Kalashnikov, Dmitry and Levine, Sergey and Lu, Yao and Parada, Carolina and Rao, Kanishka and Sermanet, Pierre and Toshev, Alexander T and Vanhoucke, Vincent and Xia, Fei and Xiao, Ted and Xu, Peng and Yan, Mengyuan and Brown, Noah and Ahn, Michael and Cortes, Omar and Sievers, Nicolas and Tan, Clayton and Xu, Sichun and Reyes, Diego and Rettinghouse, Jarek and Quiambao, Jornell and Pastor, Peter and Luu, Linda and Lee, Kuang-Huei and Kuang, Yuheng and Jesmonth, Sally and Joshi, Nikhil J. and Jeffrey, Kyle and Ruano, Rosario Jauregui and Hsu, Jasmine and Gopalakrishnan, Keerthana and David, Byron and Zeng, Andy and Fu, Chuyuan Kelly},
  booktitle={Proceedings of Conference on Robot Learning},
  pages={287--318},
  year={2023}
}

@inproceedings{liang2023code,
  title={Code as policies: Language model programs for embodied control},
  author={Liang, Jacky and Huang, Wenlong and Xia, Fei and Xu, Peng and Hausman, Karol and Ichter, Brian and Florence, Pete and Zeng, Andy},
  booktitle={Proceedings of IEEE International Conference on Robotics and Automation (ICRA)},
  pages={9493--9500},
  year={2023}
}

@inproceedings{ren2023robots,
  title={Robots That Ask For Help: Uncertainty Alignment for Large Language Model Planners},
  author={Ren, Allen Z. and Dixit, Anushri and Bodrova, Alexandra and Singh, Sumeet and Tu, Stephen and Brown, Noah and Xu, Peng and Takayama, Leila and Xia, Fei and Varley, Jake and Xu, Zhenjia and Sadigh, Dorsa and Zeng, Andy and Majumdar, Anirudha},
  booktitle={Proceedings of the Conference on Robot Learning (CoRL)},
  year={2023},
  pages = {661--682},
}

@inproceedings{wu2023tidybot,
  title={TidyBot: Personalized robot assistance with large language models},
  author={Wu, Jimmy and Antonova, Rika and Kan, Adam and Lepert, Marion and Zeng, Andy and Song, Shuran and Bohg, Jeannette and Rusinkiewicz, Szymon and Funkhouser, Thomas},
  booktitle={Proceedings of IEEE/RSJ International Conference on Intelligent Robots and Systems (IROS)},
  pages={3546--3553},
  year={2023}
}

@inproceedings{driess2023palm,
  author = {Driess, Danny and Xia, Fei and Sajjadi, Mehdi S. M. and Lynch, Corey and Chowdhery, Aakanksha and Ichter, Brian and Wahid, Ayzaan and Tompson, Jonathan and Vuong, Quan and Yu, Tianhe and Huang, Wenlong and Chebotar, Yevgen and Sermanet, Pierre and Duckworth, Daniel and Levine, Sergey and Vanhoucke, Vincent and Hausman, Karol and Toussaint, Marc and Greff, Klaus and Zeng, Andy and Mordatch, Igor and Florence, Pete},
  title = {PaLM-E: an embodied multimodal language model},
  year = {2023},
  pages={8469--8488},
  booktitle = {Proceedings of the International Conference on Machine Learning},
}

@InProceedings{brohan2023rt,
  title = {{RT-2}: Vision-Language-Action Models Transfer Web Knowledge to Robotic Control},
  author = {Zitkovich, Brianna and Yu, Tianhe and Xu, Sichun and Xu, Peng and Xiao, Ted and Xia, Fei and Wu, Jialin and Wohlhart, Paul and Welker, Stefan and Wahid, Ayzaan and Vuong, Quan and Vanhoucke, Vincent and Tran, Huong and Soricut, Radu and Singh, Anikait and Singh, Jaspiar and Sermanet, Pierre and Sanketi, Pannag R. and Salazar, Grecia and Ryoo, Michael S. and Reymann, Krista and Rao, Kanishka and Pertsch, Karl and Mordatch, Igor and Michalewski, Henryk and Lu, Yao and Levine, Sergey and Lee, Lisa and Lee, Tsang-Wei Edward and Leal, Isabel and Kuang, Yuheng and Kalashnikov, Dmitry and Julian, Ryan and Joshi, Nikhil J. and Irpan, Alex and Ichter, Brian and Hsu, Jasmine and Herzog, Alexander and Hausman, Karol and Gopalakrishnan, Keerthana and Fu, Chuyuan and Florence, Pete and Finn, Chelsea and Dubey, Kumar Avinava and Driess, Danny and Ding, Tianli and Choromanski, Krzysztof Marcin and Chen, Xi and Chebotar, Yevgen and Carbajal, Justice and Brown, Noah and Brohan, Anthony and Arenas, Montserrat Gonzalez and Han, Kehang},
  booktitle = {Proceedings of Conference on Robot Learning},
  pages = {2165--2183},
  year = {2023},
}

@inproceedings{huang2022inner,
  title={Inner monologue: Embodied reasoning through planning with language models},
  author={Huang, Wenlong and Abbeel, Pieter and Pathak, Deepak and Mordatch, Igor},
  booktitle={Proceedings of Conference on Robot Learning},
  pages={1769--1782},
  year={2023}
}

@inproceedings{shinn2024reflexion,
  author = {Shinn, Noah and Cassano, Federico and Gopinath, Ashwin and Narasimhan, Karthik and Yao, Shunyu},
  title = {Reflexion: language agents with verbal reinforcement learning},
  year = {2023},
  booktitle = {Proceedings of the International Conference on Neural Information Processing Systems},
  pages = {8634--8652},
}

@inproceedings{huang2023voxposer,
  title={{VoxPoser}: Composable 3d value maps for robotic manipulation with language models},
  author={Huang, Wenlong and Wang, Chen and Zhang, Ruohan and Li, Yunzhu and Wu, Jiajun and Fei-Fei, Li},
  booktitle={Proceedings of Conference on Robot Learning (CoRL)},
  pages={540--562},
  year={2023}
}

@inproceedings{yu2023language,
  title={Language to rewards for robotic skill synthesis},
  author={Yu, Wenhao and Gileadi, Nimrod and Fu, Chuyuan and Kirmani, Sean and Lee, Kuang-Huei and Arenas, Montserrat Gonzalez and Chiang, Hao-Tien Lewis and Erez, Tom and Hasenclever, Leonard and Humplik, Jan and Ichter, Brian and Xiao, Ted and Xu, Peng and Zeng, Andy and Zhang, Tingnan and Heess, Nicolas and Sadigh, Dorsa and Tan, Jie and Tassa, Yuval and Xia, Fei},
  booktitle={Proceedings of Conference on Robot Learning (CoRL)},
  pages={374--404},
  year={2023}
}

@techreport{EC2021Industry5,
  author      = {{European Commission}},
  title       = {Industry 5.0: Towards a sustainable, human-centric and resilient European industry},
  institution = {Directorate-General for Research and Innovation},
  year        = {2021},
  url         = {https://data.europa.eu/doi/10.2777/308407}
}

@article{liu2026ar,
  title = {AR-assisted human-robot collaborative assembly system: Integrating visual language model and deep reinforcement learning for task planning and seamless interactive guidance},
  journal = {Journal of Manufacturing Systems},
  volume = {84},
  pages = {40--67},
  year = {2026},
  author = {Changchun Liu and Dunbing Tang and Haihua Zhu and Zequn Zhang and Liping Wang and Qingwei Nie}
}

@article{cai2026llm,
  title={LLM-enhanced embodied multi-agent manufacturing system: A novel self-organizing production paradigm for embodied perception, embodied analysis and embodied decision},
  author={Changchun Liu and Dunbing Tang and Haihua Zhu and Liping Wang and Qixiang Cai and Qingwei Nie},
  journal={Journal of Manufacturing Systems},
  volume={84},
  pages={357--382},
  year={2026}
}

@article{liu2026from,
  title={From insight to action: Embodied multi-agent system integrating vision language model for digital twin-assisted human-robot collaborative assembly},
  author={Liu, Changchun and Qian, Y. and Tang, Dunbing and Zhu, Haihua and Pang, Jianhui and Cai, Qixiang},
  journal={Journal of Manufacturing Systems},
  volume={85},
  pages={531--556},
  year={2026}
}

@article{fan2025vision,
  title={Vision-language model-based human-robot collaboration for smart manufacturing: A state-of-the-art survey},
  author={Fan, Junming and Yin, Yue and Wang, Tian and Dong, Wenhang and Zheng, Pai and Wang, Lihui},
  journal={Frontiers of Engineering Management},
  volume={12},
  number={1},
  pages={177--200},
  year={2025}
}

@article{fan2024vision,
  title={A vision-language-guided robotic action planning approach for ambiguity mitigation in human--robot collaborative manufacturing},
  author={Fan, Junming and Zheng, Pai},
  journal={Journal of Manufacturing Systems},
  volume={74},
  pages={1009--1018},
  year={2024}
}

@article{wu2025h2r,
  title={{H2R Bridge}: Transferring vision-language models to few-shot intention meta-perception in human robot collaboration},
  author={Wu, Duidi and Zhao, Qianyou and Fan, Junming and Qi, Jin and Zheng, Pai and Hu, Jie},
  journal={Journal of Manufacturing Systems},
  volume={80},
  pages={524--535},
  year={2025}
}

@inproceedings{kuffner2000rrtconnect,
  title={{RRT-Connect}: An Efficient Approach to Single-Query Path Planning},
  author={Kuffner, James J. and LaValle, Steven M.},
  booktitle={Proceedings of IEEE International Conference on Robotics and Automation (ICRA)},
  pages={995--1001},
  year={2000},
}

@article{chitta2012moveit,
  title={{MoveIt!}},
  author={Chitta, Sachin and Sucan, Ioan and Cousins, Steve},
  journal={IEEE Robotics \& Automation Magazine},
  volume={19},
  number={1},
  pages={18--19},
  year={2012},
}

\end{document}